\documentclass[lettersize,journal]{IEEEtran}
\usepackage{amsmath,amsfonts}
\usepackage{algorithmic}
\usepackage{algorithm}
\usepackage{array}
\usepackage[caption=false,font=normalsize,labelfont=sf,textfont=sf]{subfig}
\usepackage{textcomp}
\usepackage{stfloats}
\usepackage{url}
\usepackage{verbatim}
\usepackage{graphicx}
\usepackage{cite}
\usepackage{hyperref}
\usepackage{makecell}
\makeatletter
\let\comment\@undefined
\makeatother
\usepackage[final ]{changes}
\begin{document}

\title{Visual Perception Engine: Fast and Flexible Multi-Head Inference for Robotic Vision Tasks}

\author{Jakub Łucki$^{1,2}$, 
Jonathan Becktor$^{1}$, 
Georgios Georgakis$^{1}$, 
Rob Royce$^{1}$ and 
Shehryar Khattak$^{1}$ 
\thanks{$^{1}$ Jet Propulsion Laboratory, California Institute of Technology, Pasadena, CA, USA.}
\thanks{$^{2}$ Swiss Federal Institute of Technology (ETH Zürich), Zürich, Switzerland.}
\thanks{The research was carried out at the Jet Propulsion Laboratory, California Institute of \replaced{Technology}{sTechnology}, under a contract with the National Aeronautics and Space Administration (80NM0018D0004).}
\thanks{\copyright 2025. All rights reserved.}}



\maketitle
\begin{abstract}
Deploying multiple machine learning models on resource-constrained robotic platforms for different perception tasks often results in redundant computations and complex integration challenges. In response, this work presents \added{the }Visual Perception Engine (VPEngine), a modular framework designed to enable efficient GPU usage for visual multitasking while maintaining extensibility and developer accessibility. Our framework architecture leverages a shared foundation model backbone that extracts image representations, which are efficiently shared, without any unnecessary GPU-CPU memory transfers, across multiple specialized task-specific model heads running in parallel. \deleted{This design eliminates the computational redundancy inherent in feature extraction component when deploying traditional sequential models while enabling dynamic task prioritization based on application demands.} We demonstrate our framework's capabilities through an example implementation using DINOv2 as the foundation model \replaced{with multiple task heads for depth estimation, object detection, and semantic segmentation}{with multiple task (depth, object detection and semantic segmentation) heads}, achieving up to \added{a }3x speedup compared to sequential execution. \deleted{Building on CUDA Multi-Process Service (MPS), VPEngine offers efficient GPU utilization and maintains a constant memory footprint while allowing per-task inference frequencies to be adjusted dynamically during runtime.} The framework is written in Python and is open source with ROS2 C++ (Humble) bindings for ease of use by the robotics community across diverse robotic platforms. Our example implementation demonstrates end-to-end real-time performance at $\geq$50 Hz on \added{an }NVIDIA Jetson \replaced{AGX Orin}{Orin AGX} for TensorRT\added{-}optimized models.~\looseness=-1 
\end{abstract}


\vspace{-1ex}
\section{Introduction}
\IEEEPARstart{V}{isual} perception is a cornerstone of autonomous robotic \replaced{systems}{system applications}, enabling mobile machines to interpret complex environments \cite{thrun2005probabilistic}, make informed decisions under uncertainty \cite{siciliano2016springer}, and interact seamlessly within dynamic real-world scenarios \cite{corke2017robotics}. The ability to extract meaningful information from visual data directly impacts a robot's ability for navigation, manipulation, obstacle avoidance, and human-robot interaction, making robust perception systems essential for reliable autonomous operations.

Recent advances in computer vision have \replaced{driven}{witnessed} a paradigm shift toward visual foundation models that can serve multiple perception tasks through shared representations. Vision transformers trained on internet\added{-}scale data\replaced{, including}{ like} DINOv2 \cite{oquab2024dinov}, \replaced{SigLIP 2}{SAM} \cite{tschannen2025siglip2}, and CLIP \cite{radford2021learning}\added{,} have demonstrated remarkable capabilities \deleted{in replacing specialized solutions }across diverse applications. \deleted{For instance, DINOv2 extracts image representations that generalize across domain and as such can be used for a variety of vision tasks without any finetuning.}

This convergence toward \replaced{general}{unified} architectures suggests that leveraging one large network for multiple perception tasks should be the natural progression for robotic systems. However, current robotic perception frameworks remain inadequate or inefficient in their implementation. For example, NVIDIA Isaac ROS \cite{nvidia_isaac_ros}, while providing \replaced{highly}{excellent} optimized perception capabilities, deploys separate specialized models for each task rather than utilizing a shared foundation model backbone. Isaac ROS, among others, offers individual nodes for stereo depth estimation (ESS), object detection (YOLOv8), and semantic segmentation (\replaced{SegFormer}{Segformer}), each running independently. This approach, while functionally robust, perpetuates computational redundancy, as \replaced{general}{unified} vision features can be used for multipurpose downstream tasks~\cite{oquab2024dinov}.
\begin{figure}[!t]
\centering
\includegraphics[width=\columnwidth]{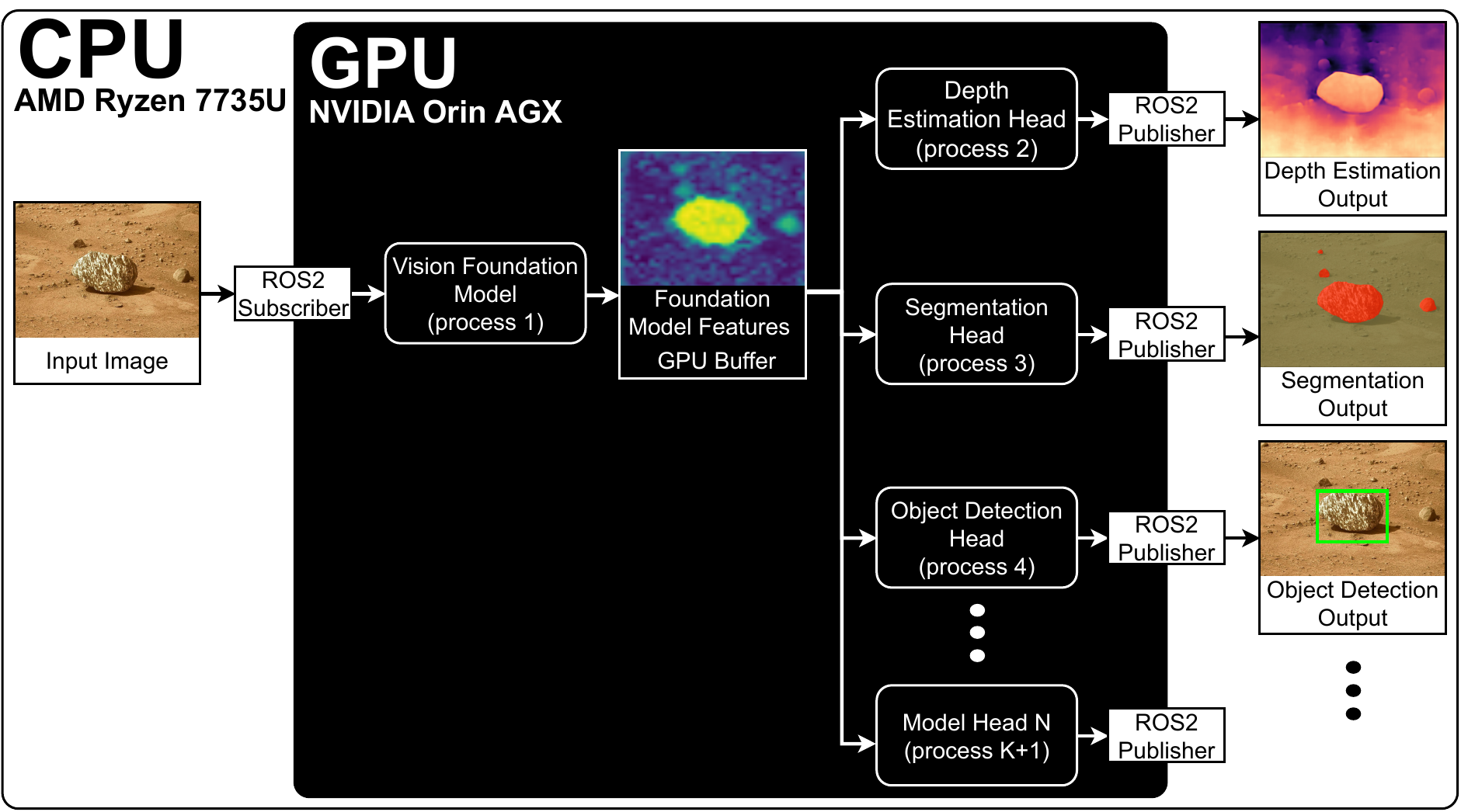}
\vspace{-4ex}
\caption{Example implementation use case of the proposed Visual Perception Engine (\replaced{VPEngine}{VPE}) for three downstream robotic vision tasks. \replaced{VPEngine}{VPE} enables \replaced{zero-copy sharing of}{without copying sharing of} visual foundation features from \added{a }TensorRT-optimized backbone with multiple independently TensorRT-optimized (depth estimation and semantic segmentation) or PyTorch (object detection) task heads. The CPU and GPU tasks are represented by white and black boxes, respectively.}
\label{fig:perception_engine_overview}
\vspace{-3ex}
\end{figure}

To address these limitations, we introduce \added{the }Visual Perception Engine (VPEngine), a framework specifically designed to harness the power of visual foundation models for robotic multitasking efficiently. VPEngine uses highly optimized TensorRT models\footnote{However, it supports native PyTorch as well, since it is not always feasible to convert a model to TensorRT.}\replaced{, which run}{, running} in separate processes\replaced{ and, when combined with CUDA MPS, result}{, which combined with CUDA MPS results} in \replaced{low-latency}{a very fast} inference and high GPU utilization. \added{Our primary contributions are an open-source}~\footnote{\url{https://github.com/nasa-jpl/visual-perception-engine}} \replaced{modular framework written in Python enabling fast prototyping and easy extensibility for efficient multitask ML inference; a custom inter-process communication system enabling \textit{by-reference} GPU memory transfers across processes, which are not supported on Jetson devices. This framework will benefit most robots requiring a variety of visual insights, such as planetary rovers or field robots that need traversability estimation, semantic terrain labels, and object/landmark detection.}
{Furthermore, we devised custom Inter-Process Communication structures using CUDA API to enable \textit{by reference} GPU memory transfers across processes, which are not supported on Jetson devices. As a consequence, the latency of sharing image representations between the backbone and model heads is negligible. The open-source framework is written in Python, heavily leverages abstract classes and is highly modularized which enables fast prototyping and fosters extendability. A high-level overview of the system is shown in Figure.}



\vspace{-1ex}
\section{Design Requirements}
To address the challenges of efficient visual multitasking in resource-constrained robotic systems, we establish the following design requirements that guide VPEngine\deleted{s}'s architecture:

\textbf{R1: Fast Inference} - Perception \deleted{loop} latency can be the upper bound on the robot's reaction speed to environment changes.\footnote{\added{For high-speed robotic platforms traveling at 2--5 m/s, a robot travels 4--25 cm during a 20--50 ms perception delay. We therefore target a perception latency on the order of one 30 Hz camera frame ($\approx$33 ms), while noting that there is no universal latency threshold because the exact budget depends on robot speed, stopping distance, and controller design.}}\deleted{Furthermore, if desired model inference frequencies are fixed, efficiency allows for GPU availability for other tasks.}

\textbf{R2: Memory Predictability} - Autonomous systems must operate reliably for extended periods without memory-related failures that could compromise mission success.
    
    
\textbf{R3: Extendability and Flexibility} - Robotic applications vary in their perception needs, requiring flexible architectures that accommodate diverse task combinations.~\looseness=-1
    
\textbf{R4: Dynamic Task Prioritization} - Environmental conditions during robot missions require adaptive perception systems that can prioritize critical tasks based on current context.~\looseness=-1

\vspace{-1ex}
\section{System architecture}
\begin{figure*}[!t]
\centering
\includegraphics[width=\textwidth]{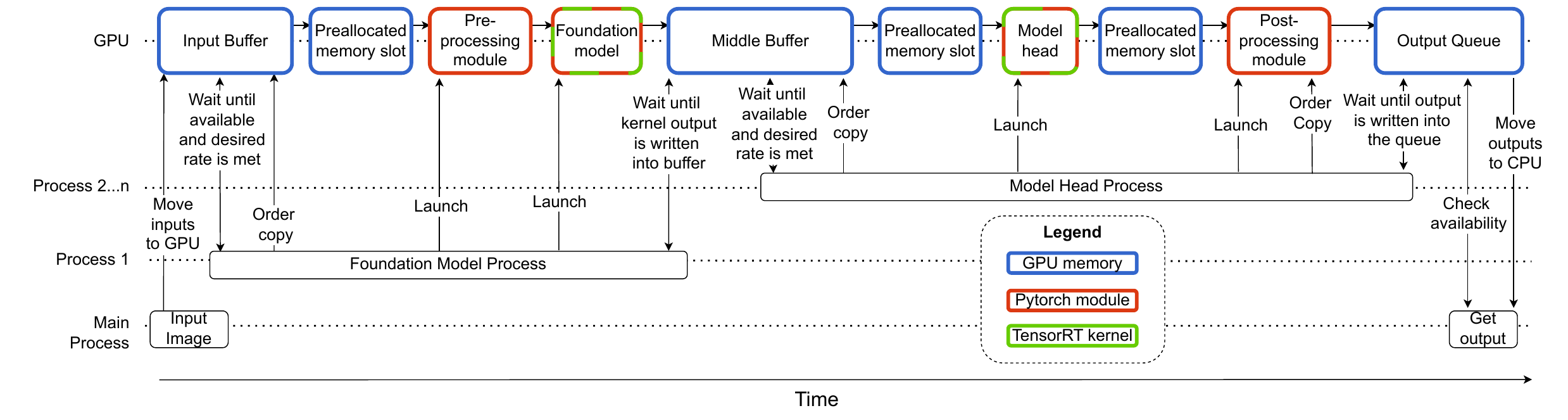}
\vspace{-5ex}
\caption{Architecture overview of the VPE showing interaction\added{s} between CPU and GPU computations. The figure \deleted{also }highlights \added{how }visual foundation model feature\added{s remain} in GPU memory and \replaced{are shared efficiently}{efficiently sharing} with \replaced{PyTorch}{Pytorch} modules and \replaced{TensorRT-optimized}{TensorRT optimized} kernels \replaced{while minimizing copies}{minimizing the number of copies}.}
\label{fig:perception_engine_detail}
\vspace{-3ex}
\end{figure*}
VPEngine consists of a foundation model (FM) serving as the computational core that provides rich visual representations, which are then efficiently shared and processed by an arbitrary number of lightweight task-specific model heads. \added{A high-level overview of the system is shown in} Figure~\ref{fig:perception_engine_overview}. Compared to independent task-specific models, this architecture eliminates redundant \added{early-stage} feature extraction \deleted{in the earlier stage of each network} by sharing \deleted{of} FM visual features among tasks in a \replaced{memory-efficient}{memory efficient} way without creating feature copies. A detailed system overview is presented in Figure~\ref{fig:perception_engine_detail}.

\vspace{-2ex}
\subsection{Modular Architecture}
VPEngine implements two distinct module types to achieve efficient visual multitasking:

\subsubsection{Foundation Module} 
The foundation module serves as the shared backbone for all perception tasks, addressing \textbf{R1} by centralizing the extraction of features. It consists of four main components: a) an input buffer that receives raw sensor data, b) pre-processing transformations that normalize inputs for the foundation model, c) the foundation model to extract visual features, and d) a middle buffer that stores computed features for sharing among task heads.

The foundation module operates continuously, processing incoming images and maintaining a constant buffer of feature representations. All computations remain strictly on \added{the }GPU to minimize memory transfer overhead.

\subsubsection{Head Module}
Task-specific head modules take FM features as input and perform specialized perception tasks, directly supporting \textbf{R3}. Each head module contains a) an interface to asynchronously \replaced{access}{accesses} the middle feature buffer, b) the lightweight task-specific model, c) postprocessing transformations to format outputs for downstream consumers, and d) an output buffer that delivers results to external systems, e.g., a ROS2 publisher. \replaced{Head}{Each head} modules operate independently, allowing each task head to execute at variable frequency based on the application requirements.


\subsubsection{Transforms}
Pre-processing and postprocessing transforms serve as adapters between user data and the core of the engine, accommodating varying tensor shapes\deleted{,} and data types. These lightweight operations ensure compatibility between foundation models and task heads without requiring modifications to the core neural networks.

Transform operations are implemented as simple tensor manipulations (reshaping, normalization, type conversion) that execute efficiently on GPU without significant computational overhead. This abstraction layer enables seamless integration of models with different input/output specifications.

\subsubsection{Module separation}
Each model component executes in a separate process to maximize GPU utilization through CUDA MPS, addressing both \textbf{R1} and \textbf{R3}. To utilize CUDA MPS, a standard CUDA MPS server is configured on the deployment device, after which all CUDA-enabled processes automatically leverage shared GPU resources. An additional benefit of CUDA MPS is that it allocates only one CUDA context for all processes\cite{nvidia_mps}.

This multi-process design enables true parallelization when individual models cannot fully saturate GPU resources, a common scenario when portions of models execute in PyTorch rather than optimized TensorRT kernels. The framework also provides fault isolation where \added{a }failure in one task head does not compromise the entire perception pipeline.

While multi-process deployment introduces PyTorch initialization overhead across processes, the benefits of better encapsulation, fault isolation, and parallel execution through CUDA MPS significantly outweigh these costs in practice. \added{Lastly, although the MPS server introduces an additional point of failure, its failure does not crash VPEngine but only slows it down. For safety-critical or long-duration deployments, we recommend supervising the MPS daemon with a watchdog or service manager.}

\subsubsection{Decentralization}
Each module operates independently with minimal central coordination, supporting \textbf{R4}. Modules execute as soon as input data becomes available, subject only to their configured frequency limits. \replaced{Since the modules never synchronize, the whole system is lossy in the worst case. This prioritizes result freshness, which we determined to be more important in dynamic environments.}{Runtime parameters such as execution frequency can be adjusted through the main process coordination system, but modules maintain autonomous operation for maximum responsiveness.}~\looseness=-1

This decentralized approach ensures that high-priority tasks are not blocked by slower components and enables adaptive resource allocation based on changing environmental demands.

\vspace{-1ex}
\subsection{Memory Management}
VPEngine implements `by\added{-}reference' GPU memory sharing, directly addressing \textbf{R2} and contributing to \textbf{R1}. This is not \replaced{supported}{suported} on Jetson platforms by default\replaced{; hence,}{ hence} we created our own Inter-Process Communication structures leveraging \added{the }low-level CUDA API, in particular \deleted{function} \texttt{cuMemImportFromShareableHandle}. As a consequence, foundation model outputs remain GPU-resident, with task heads accessing shared memory regions directly via CUDA pointers. This design eliminates costly GPU-CPU-GPU memory transfers that create overhead for traditional multi-model pipelines.

Memory copies only occur when removing elements from queues/buffers into the pre-allocated processing slots. This copy operation is necessary to release queue slots for new data; otherwise, slots would remain locked until model heads complete processing, potentially causing pipeline stalls.

\replaced{Each queue and buffer preallocates the required GPU memory at system initialization, which is then reused during steady-state operation. This ensures a constant memory footprint and prevents runtime memory allocation failures such as OOMs (\textbf{R2}). Since startup is the only phase when preallocation happens, the risk of memory fragmentation is low; however, for long-duration deployments, we recommend runtime memory monitoring.}{Memory allocation follows a static pattern established during initialization, with all queue and buffer sizes specified in the configuration file (\textit{config.json}). Each queue and buffer pre-allocates required GPU memory during system startup, ensuring constant memory footprint during operation and preventing runtime allocation failures that could compromise system stability (\textbf{R2}).}

\vspace{-1ex}
\subsection{Data Flow}
Inter-process communication utilizes two data structures:
\subsubsection{GPU Queues} FIFO structures for ordered data transfer between FM and task heads. Queues signal overflow conditions when consumers cannot keep pace with producers.~\looseness=-1
\subsubsection{GPU Buffers } Circular buffers that overwrite \added{the }oldest data when full, ensuring that the models always access the most recent tensors. This design prioritizes temporal relevance over processing completeness.

All tensors are organized as labeled dictionaries, enabling flexible data routing. Task heads can selectively consume relevant FM features from the complete output dictionary without index-based coupling.
\vspace{-1ex}
\subsection{Dynamic Frequency Control}
Each task head supports runtime frequency adjustment through a control interface. This enables adaptive perception where task execution rates adjust based on environmental changes, computational load, or mission requirements (\textbf{R4}). For example, \replaced{the frequency of obstacle detection}{obstacle detection} might increase in dense environments while \replaced{the semantic segmentation rate is reduced}{reducing semantic segmentation rate} to balance computational resources for robotic tasks.

\vspace{-1ex}
\subsection{Model Optimization}
The system leverages NVIDIA TensorRT for inference optimization, directly supporting \textbf{R1} and \textbf{R2}. TensorRT compiles neural networks into optimized execution engines that maximize GPU throughput through kernel fusion, precision optimization, and memory layout optimization. This compilation process is particularly crucial for foundation models where inference latency directly impacts the entire pipeline's performance. \replaced{VPEngine}{Please note in our framework} allows \deleted{for} \replaced{the FM and task heads to be independently compiled with TensorRT}{FM and model heads to be independently TensorRT compiled} while \replaced{still maintaining}{maintaining} efficient FM feature sharing between them.

\deleted{
A unified model registry manages all foundation models and task heads, including version control, metadata tracking, and deployment configuration. The registry supports automatic TensorRT compilation, model validation, and deployment verification across different hardware targets. VPEngine supports both TensorRT-optimized engines for maximum efficiency and native PyTorch models for development flexibility.}

\vspace{-1ex}
\section{Example implementation}\label{sec:example_implementation}
In our example implementation, we use three model heads: DepthAnythingV2 (DAV2) \cite{yang2024depth} \deleted{responsible }for monocular depth estimation\replaced{, compiled with TensorRT using}{ (we used} optimized code from \cite{spacewalk01_depth_anything_tensorrt_2024}\deleted{) compiled with TensorRT}, \added{a }linear layer for semantic segmentation \cite{oquab2024dinov} compiled with TensorRT, and a custom \replaced{Faster R-CNN}{FasterRCNN} \replaced{object detection head (in pure PyTorch)}{head object detection}. To get vision foundation model representations\added{,} we use \added{the }ViT-S version of DINOv2 \cite{oquab2024dinov}. We configure it to output final features along with representations computed at three evenly spaced intermediate layers into the middle buffer, since they are required by \added{the }DAV2 model head.

We test our example implementation on a stream of 30000 images of $1920\times1080$ resolution, and it maintains~\footnote{Due to \added{the }freshness-oriented design and stochastic nature of the engine\added{,} minimal losses are unavoidable, especially when using models not compiled with TensorRT. However, even then they account for $\approx0.02\%$ of all frames.} a 30 Hz throughput with median inference times of $20$\textit{ms}, $18$\textit{ms}, and $32$\textit{ms} for the depth estimation, semantic segmentation, and object detection heads, respectively, on \replaced{an NVIDIA Jetson}{a Nvidia} Orin AGX 64GB. It is important to note that the object detection head is not compiled with TensorRT, \replaced{which explains}{hence} the larger inference time. Throughout the test, GPU memory usage was constant at around $1.5$GB. Using VPEngine, the combined number of parameters \replaced{in}{of} our example implementation is 27M\footnote{Using our custom object detection head and the others provided with~\cite{oquab2024dinov}.}, which \added{is} 69\% lower than if we used three independent models~\footnote{Using DINOv2 models for semantic segmentation and depth estimation, and~\href{https://docs.pytorch.org/vision/main/models/generated/torchvision.models.detection.fasterrcnn_resnet50_fpn.html}{\replaced{Faster R-CNN}{FasterRCNN} implementation in \replaced{PyTorch}{Pytorch}}.}





\vspace{-1ex}
\section{Benchmarks}
To validate the design choices and quantify performance, we conducted two experiments. First, we compared our single multi-head model design to a series of independent, task-specific models. Second, we quantified the performance difference between a multiprocessing architecture and a single-process design. Both experiments use processing latency and peak GPU memory usage\added{, as well as average total CPU utilization,} as primary metrics. To measure latency, we record the time from inserting an input image as a CPU array to the moment all heads or baseline models return CPU tensors\footnote{This is an approximation, or rather a lower bound on the engine's throughput, because it limits the engine's ability to parallelize across inputs.}. All experiments were conducted under identical conditions using an NVIDIA Jetson AGX Orin 64GB.

\vspace{-2ex}
\subsection{\replaced{Single}{Unified} Backbone vs. Independent Models}

This experiment highlights the benefit of using a \replaced{single}{unified} backbone for multiple tasks instead of separate, full models. We compared VPEngine against two setups:
\begin{itemize}
    \item \textbf{DepthAnythingV2}: A sequence of complete, independent DepthAnythingV2 models (each with its own backbone), compiled with TensorRT.
    \item \textbf{Sequential Heads}: A single, shared DINOv2 backbone followed by a sequence of TensorRT-optimized heads running one after another in the same process.
\end{itemize}

\begin{figure}[t]
\centering
\includegraphics[width=\linewidth]{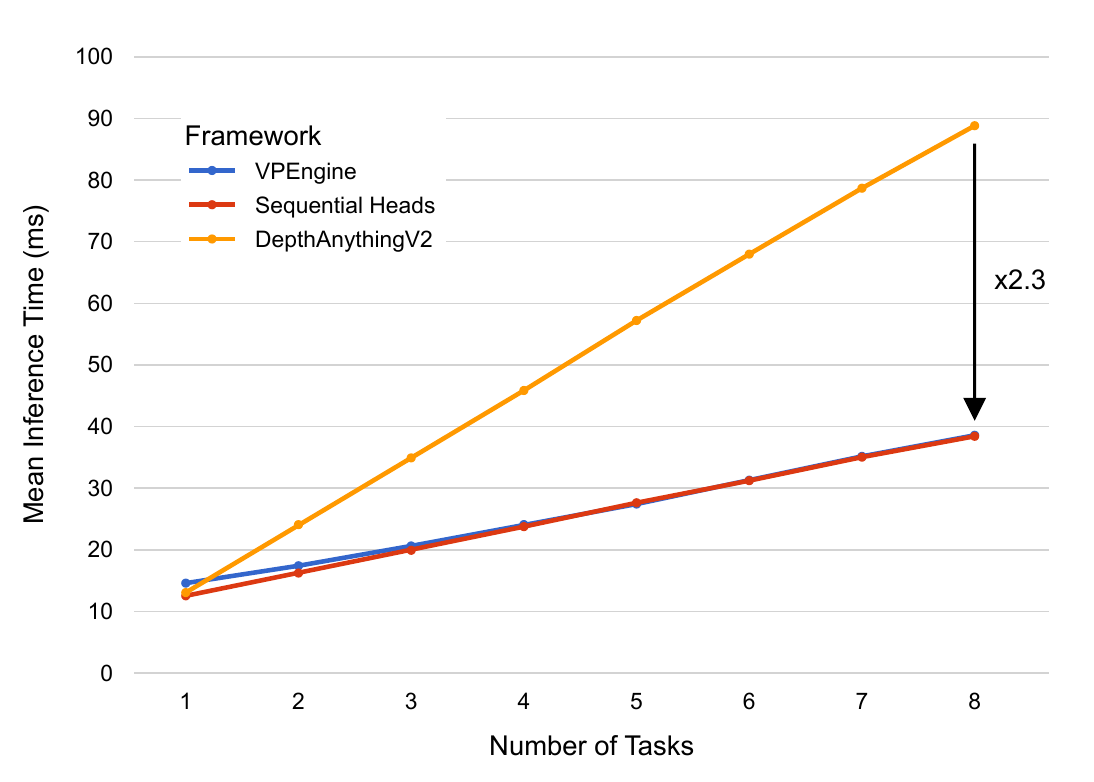}
\vspace{-5ex}
\caption{Mean inference time across different \replaced{numbers}{number} of tasks over 426 test images (of size $1920\times1080$). The arrow indicates \added{a }$2.3\times$ speed gain of a single backbone with a sequence of 8 TensorRT heads over 8 sequential DepthAnythingV2 models.}
\label{fig:multihead_design}
\vspace{-3ex}
\end{figure}

As shown in Figure~\ref{fig:multihead_design}, using a shared foundation model provides a significant speedup of up to \textbf{2.3$\times$} compared to running multiple independent models. VPEngine performance is nearly identical to the ``Sequential Heads'' setup. The TensorRT-optimized kernels in this test already achieve nearly 100\% GPU utilization, leaving little room for parallel processing to provide additional benefits. VPEngine introduces a negligible overhead of only about 2\textit{ms}, which is due to the communication between processes. Regarding memory, the full DepthAnythingV2 model uses an additional 96MB of GPU memory for each new task, the Sequential Heads baseline uses 48MB, and VPEngine uses 214MB per task. \added{With four heads, VPEngine uses approximately 438\% average total CPU utilization, compared with approximately 70\% for Sequential Heads and 65\% for DepthAnythingV2 (Fig.~\ref{fig:multihead_design_cpu}).}

\vspace{-1ex}
\subsection{Parallel vs. Sequential Processing}\label{sec:parallel_design}

\begin{figure}[t]
\centering
\includegraphics[width=\linewidth]{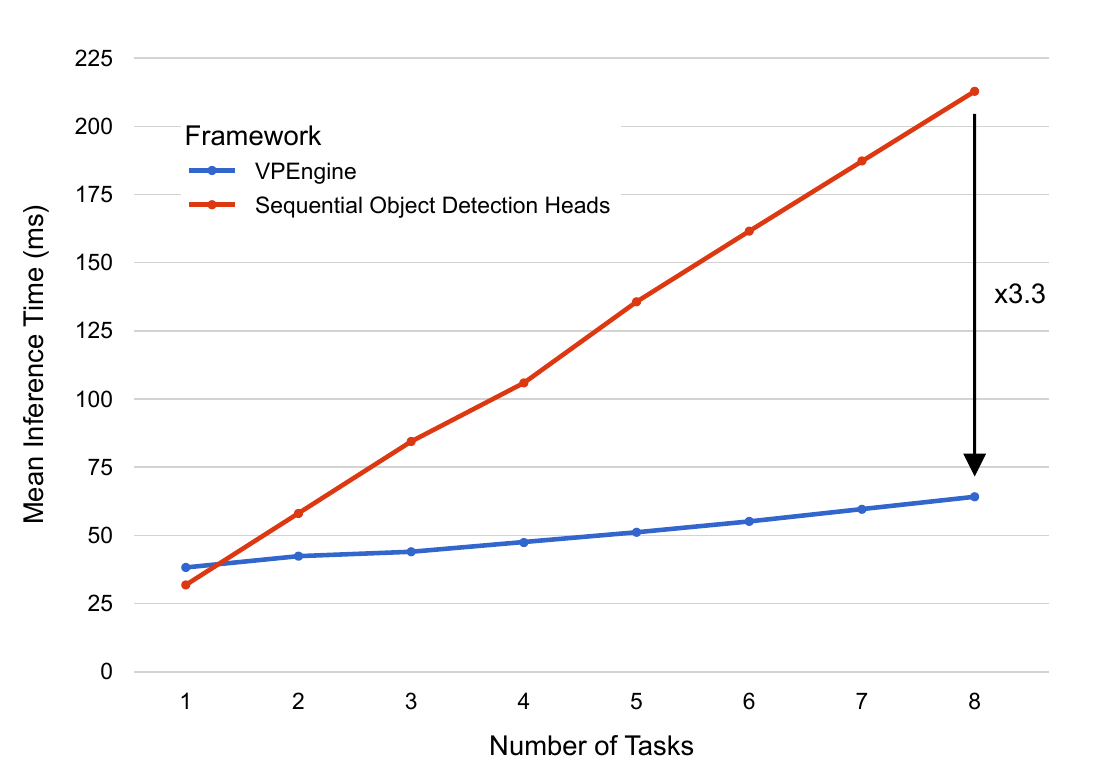}
\vspace{-5ex}
\caption{Mean inference time across different \replaced{numbers}{number} of tasks over 426 test images (of size $1920\times1080$). The arrow indicates \added{a }$3.3\times$ speed gain of VPEngine \added{running 8 PyTorch object detection heads} over a single \replaced{TensorRT}{TensoRT} foundation model with 8 PyTorch object detection heads.}
\label{fig:parallel_design}
\vspace{-3ex}
\end{figure}

This experiment quantifies the advantage of VPEngine's ability to run multiple model heads in parallel. We compared our framework against a ``Sequential Object Detection Heads'' baseline, which uses a single \added{TensorRT} backbone but runs multiple PyTorch-based object detection heads one after another. \added{In this configuration, VPEngine runs the same unoptimized PyTorch heads.} Unlike the TensorRT kernels in the previous experiment, these PyTorch heads do not fully utilize the GPU. Results in Figure~\ref{fig:parallel_design} show that VPEngine's parallel, multi-process architecture is up to \textbf{3.3$\times$ faster}. This speedup comes from running several less-optimized kernels at the same time, which improves overall GPU utilization. \added{The }significance of this design choice is highlighted by the fact that disabling CUDA MPS resulted in a 43.4\% loss in speed. This performance gain comes with an increase in memory \added{and CPU} usage\added{, both of which scale linearly with the number of tasks. See Appendix~\ref{app:parallel_design} for details.} \deleted{VPEngine requires about 295MB of GPU memory for each additional object detection head, whereas the sequential baseline only needs about 5MB. This higher memory footprint is because our multi-process design requires a separate instance of PyTorch (which uses $\approx 110$MB) to be loaded for each process. Each instance then manages its own separate block of GPU memory, unlike a single-process application that can reuse the same memory block for sequential tasks.}

\vspace{-1ex}
\section{Limitations}
A limitation of our framework is the use of multiple processes, leading to higher CPU usage. This also results in slightly higher GPU memory usage\footnote{In our benchmarks\added{,} we are using relatively small models\replaced{; hence, the extra}{, hence extra} $295$MB seems large in comparison\replaced{. However, many}{, but please note that many} models such as DINOv2-Giant use \added{an }order of magnitude more memory (4.4GB in this case).}. \added{This higher resource usage may make our framework unsuitable for highly specialized robots, such as drones with stringent resource constraints.}
\deleted{Additionally, different modules are running independently, which leads to overall stochastic nature of VPEngine. Implementing coordination between different processes, could potentially result in even higher GPU usage. We leave this to future work.}

\added{Although VPEngine can reduce perception latency, its effect on a robot's closed-loop behavior depends on the controller, robot dynamics, planner update rate, and safety margins.}




\vspace{-1ex}
\section{Conclusion}
We present VPEngine, a framework that offers substantial speedups through efficient GPU usage while still providing significant \replaced{architectural}{architecture} flexibility. For example, it speeds up execution of \replaced{PyTorch}{Pytorch} models by 3.3 times compared to running task-specific model heads sequentially.

This comes with a trade-off of higher CPU usage and slightly higher GPU memory usage. As a consequence, this might not be a \replaced{low-resource}{low resource} framework\replaced{. However, if}{, however, for example, if} a robotic stack is not using its full CPU capacity, the speedup offered may outweigh the tradeoffs.

\added{Future work could focus on better coordination between separate processes to increase GPU utilization either through a custom scheduler or by employing an inference server such as Triton~\cite{nvidia_triton_inference_server}. Alternatively, one could rewrite VPEngine to use CUDA streams instead of separate processes to reduce CPU utilization.}



 
\newpage
\bibliographystyle{IEEEtran}
\bibliography{bibliography}
%
\clearpage
\begin{appendices}

\section{Multihead\deleted{ed} design - GPU and CPU \replaced{Results}{results}}\label{app:multiheaded_design}
\begin{figure}[H]
\centering
\subfloat[Peak GPU memory.\label{fig:multihead_design_gpu}]{%
    \includegraphics[width=\linewidth]{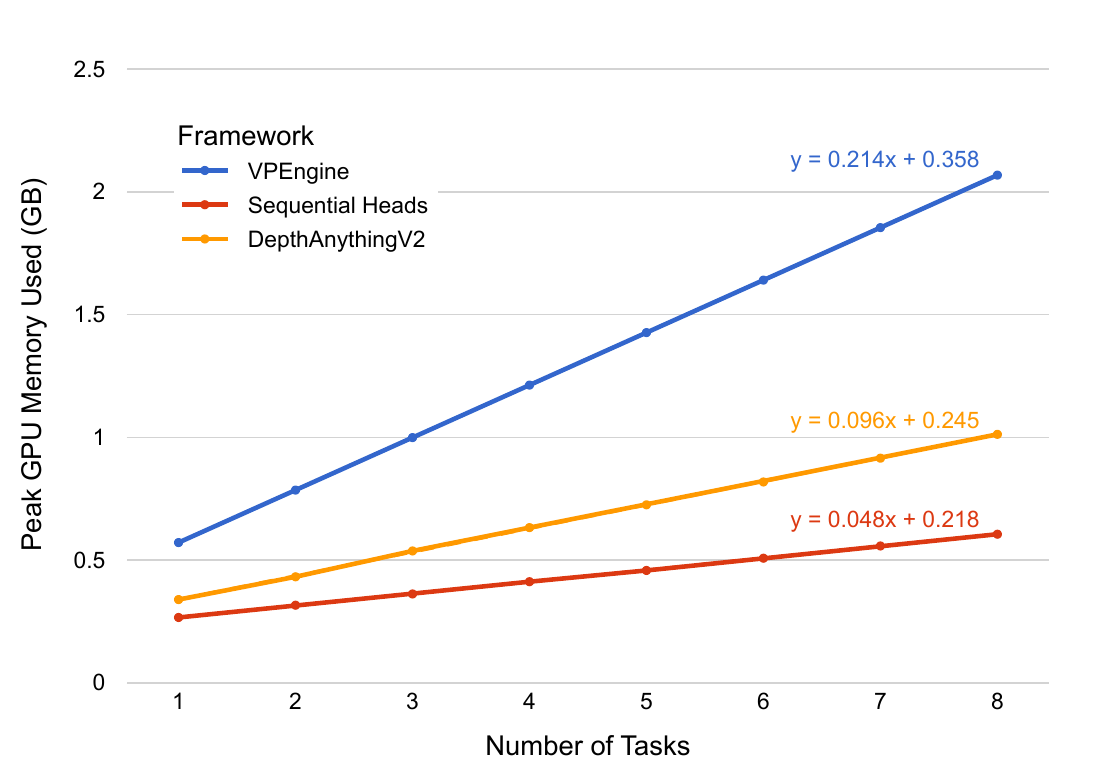}}
\par\vspace{-1ex}
\subfloat[Average total CPU utilization.\label{fig:multihead_design_cpu}]{%
    \includegraphics[width=\linewidth]{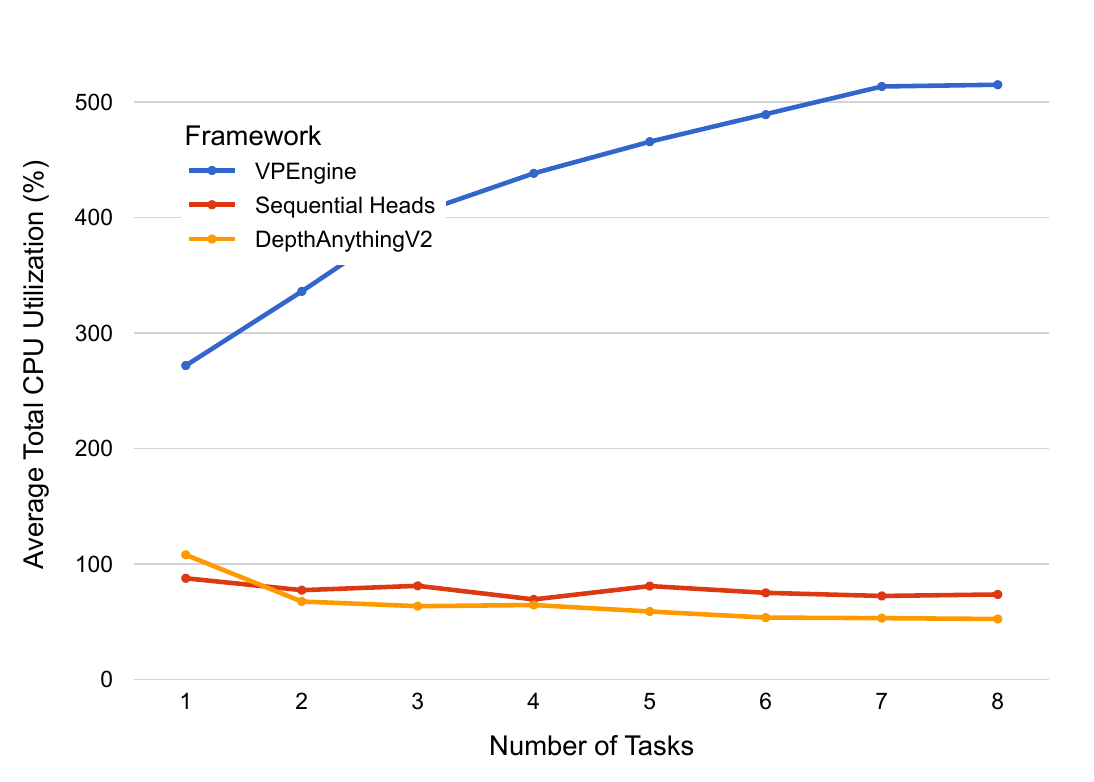}}
\caption{\replaced{Resource usage of the multihead}{Multiheaded} design \replaced{for}{resource usage across} different \replaced{numbers}{number} of tasks over 426 test images \replaced{at a resolution of}{(of size} $1920\times1080$\deleted{)}. CPU utilization is averaged over the experiment, \replaced{with}{where} 100\% \replaced{representing}{represents} full utilization of one core. GPU memory is reported as peak usage.}
\label{fig:multihead_design_resources}
\end{figure}

\section{Parallel \replaced{Design}{design} - GPU and CPU \replaced{Results}{results}}\label{app:parallel_design}
\begin{figure}[!t]
\centering
\subfloat[Peak GPU memory.\label{fig:parallel_design_gpu}]{%
    \includegraphics[width=\linewidth]{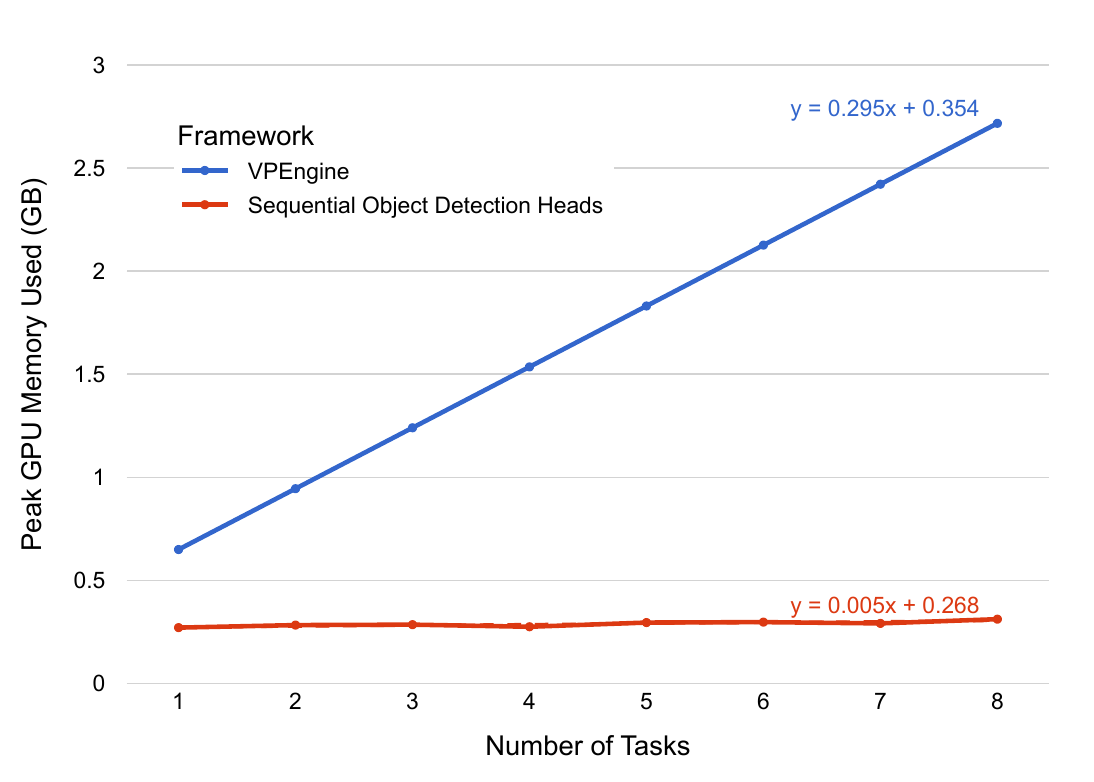}}
\par\vspace{-1ex}
\subfloat[Average total CPU utilization.\label{fig:parallel_design_cpu}]{%
    \includegraphics[width=\linewidth]{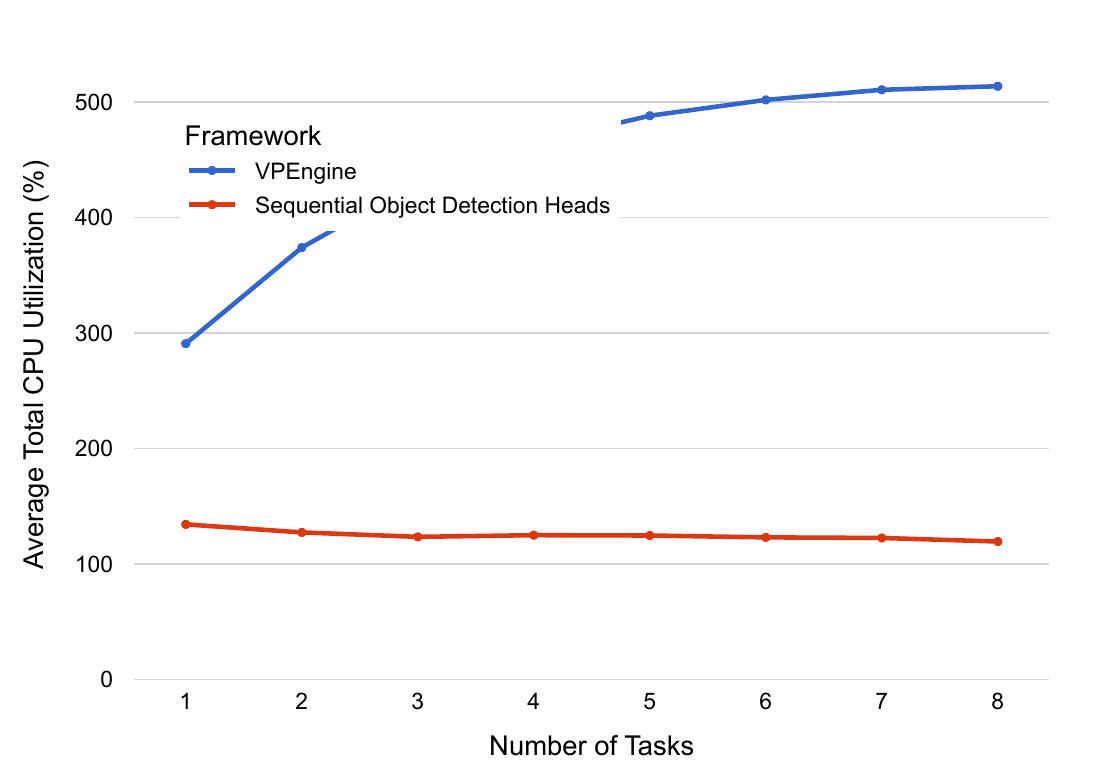}}
\caption{\replaced{Resource usage of the parallel}{Parallel} design \replaced{for}{resource usage across} different \replaced{numbers}{number} of tasks over 426 test images \replaced{at a resolution of}{(of size} $1920\times1080$\deleted{)}. CPU utilization is averaged over the experiment, \replaced{with}{where} 100\% \replaced{representing}{represents} full utilization of one core. GPU memory is reported as peak usage.}
\label{fig:parallel_design_resources}
\end{figure}

\added{VPEngine requires approximately 295 MB of GPU memory for each additional object-detection head, whereas the sequential baseline requires only approximately 5 MB. This higher memory footprint arises because our multiprocess design loads a separate instance of PyTorch, which uses approximately 110 MB, for each process. Each instance then manages its own block of GPU memory, unlike a single-process application that can reuse the same memory block for sequential tasks. The multiprocess design also increases CPU usage: with four heads, VPEngine uses approximately 465\% average total CPU utilization, compared with approximately 125\% for Sequential Object Detection Heads (Fig.~\ref{fig:parallel_design_cpu}).}

\section{CUDA MPS Ablation Results}\label{app:cuda_mps_ablation}
We \replaced{evaluate}{ablate} the \replaced{benefit of}{gain from using} CUDA MPS\deleted{ by} using the VPEngine \replaced{configuration}{setup} from Section\replaced{~\ref{sec:parallel_design}}{ \ref{sec:parallel_design}} and measuring \replaced{throughput}{the speed} with CUDA MPS first disabled and then \replaced{enabled}{enables}. \replaced{The results}{Results} are\added{ shown} in Table\added{~\ref{tab:cuda_mps_ablation}}.\deleted{\ref{tab:cuda_mps_ablation}}
\begin{table}[H]
\renewcommand{\arraystretch}{1.3}
\caption{CUDA MPS Ablation - Image\added{-}Processing Throughput}
\label{tab:cuda_mps_ablation}
\centering
\begin{tabular}{c||c|c|c}
\hline
\bfseries \makecell{Number of \\ tasks} & \bfseries \makecell{CUDA MPS \\ Disabled (Hz)} & \bfseries \makecell{CUDA MPS \\ Enabled (Hz)} & \bfseries \makecell{Relative \\ Speedup} \\
\hline\hline
4 & 15.24 & 21.23 & 39.3\% \\
\hline
8 & 8.77 & 15.52 & 77.0\% \\
\hline
\end{tabular}
\end{table}

\section{\replaced{Rate-Limiting Experiment}{Rate limiting experiment}}\label{app:rate_limiting}
We conduct an experiment to test \replaced{the engine's rate-limiting}{rate limiting} feature\added{. The setup consists} of\deleted{ our engine. Our setup involves} a stream of 300 images published at \replaced{30 Hz}{30Hz} over 10 seconds, \replaced{with}{where} all \replaced{four}{4} components of our example implementation (Section\replaced{~\ref{sec:example_implementation}}{ \ref{sec:example_implementation}})\deleted{ are} set to the same inference \replaced{rate}{\textit{rate}}. \replaced{We then measure the number of}{Ultimately, we measured how many} outputs\deleted{ were} delivered by each head. Table\replaced{~\ref{tab:rate_limiting} shows that}{ \ref{tab:rate_limiting} contains} the \replaced{rate-limiting}{results, which show that our rate limiting} feature \replaced{operates}{works} with\added{ a margin of error of} $\pm 1$ \replaced{frame}{margin of error}.
\begin{table}[H]
\renewcommand{\arraystretch}{1.3}
\caption{VPEngine \replaced{rate-limiting results}{Rate Limiting Results}. \replaced{The values}{Numbers} in \replaced{parentheses indicate the}{brackets contain} ideal \replaced{numbers}{amount} of processed frames\replaced{;}{, followed by} the \replaced{bold values show the corresponding errors}{error}.}
\label{tab:rate_limiting}
\centering
\begin{tabular}{c||c|c|c}
\hline
\bfseries \makecell{Inference Rate \\ (Hz)} & \bfseries \makecell{Semantic \\ Segmentation} & \bfseries \makecell{Object \\ Detection} & \bfseries \makecell{Depth \\ Estimation} \\
\hline\hline
5 & (50) \textbf{+1} & (50) \textbf{+1} & (50) \textbf{+1} \\
\hline
10 & (100) \textbf{+1} & (100) \textbf{+1} & (100) \textbf{+1} \\
\hline
15 & (150) \textbf{+1} & (150) \textbf{+1} & (150) \textbf{+1} \\
\hline
20 & (200) \textbf{+1} & (200) \textbf{+1} & (200) \textbf{+1} \\
\hline
25 & (250) \textbf{+1} & (250) \textbf{+1} & (250) \textbf{+1} \\
\hline
30 & (300) \textbf{=0} & (300) \textbf{-1} & (300) \textbf{=0} \\
\hline
\end{tabular}
\end{table}

\end{appendices}

\end{document}